%% file: main.tex
\documentclass[runningheads]{llncs}
\usepackage[T1]{fontenc}
\usepackage{epsfig}
\usepackage{graphicx}
\usepackage{amsmath}
\usepackage{booktabs}
\usepackage{array}
\usepackage{multirow}
\usepackage{color}
\usepackage{colortbl}
\usepackage{framed}
\usepackage{hyperref}
\usepackage{bm}
\usepackage{bbm}
\usepackage{xspace} 
\usepackage{enumitem}
\usepackage{cite}
\usepackage{xparse}
\usepackage{xcolor}
\usepackage{algorithm}
\usepackage{comment}
\usepackage{listings}
\usepackage{url}
\usepackage[symbol]{footmisc}
\usepackage{pifont}
\definecolor{Gray}{gray}{0.9}
\definecolor{LightCyan}{rgb}{0.88,0.95,1}
\definecolor{blond}{rgb}{0.98, 0.94, 0.75}

\definecolor{Gray}{gray}{0.9}
\definecolor{LightCyan}{rgb}{0.88,0.95,1}
\definecolor{blond}{rgb}{0.98, 0.94, 0.75}
\definecolor{magicmint}{rgb}{0.67, 0.94, 0.82}
\definecolor{teagreen}{rgb}{0.82, 0.94, 0.75}

\def \ie {\emph{i.e.}}
\def \eg {\emph{e.g.}}
\def \etal {\emph{et al.}}

\newcommand{\tit}[1]{\smallbreak\noindent\textbf{#1.}}

\newcommand{\cmark}{\ding{51}}%
\newcommand{\xmark}{\ding{55}}%

\newcommand{\ours}{OpenFashionCLIP\xspace}

\begin{document}
\sloppy
\title{OpenFashionCLIP:\\Vision-and-Language Contrastive Learning with Open-Source Fashion Data}
\titlerunning{Vision-and-Language Contrastive Learning with Open-Source Fashion Data}
%
\author{
Giuseppe Cartella\inst{1}\orcidID{0000-0002-5590-3253} \and
Alberto Baldrati\inst{2,3}\orcidID{0000-0002-5012-5800} \and
Davide Morelli\inst{1,3}\orcidID{0000-0001-7918-6220} \and
Marcella Cornia\inst{1}\orcidID{0000-0001-9640-9385} \and
Marco Bertini\inst{2}\orcidID{0000-0002-1364-218X} \and
Rita Cucchiara\inst{1}\orcidID{0000-0002-2239-283X}
}

\authorrunning{G. Cartella et al.}
%
\institute{
University of Modena and Reggio Emilia, Modena, Italy 
\email{\{name.surname\}@unimore.it}\and
University of Florence, Florence, Italy\\
\email{\{name.surname\}@unifi.it}\and
University of Pisa, Pisa, Italy
}
\maketitle              
\begin{abstract}
The inexorable growth of online shopping and e-commerce demands scalable and robust machine learning-based solutions to accommodate customer requirements. In the context of automatic tagging classification and multimodal retrieval, prior works either defined a low generalizable supervised learning approach or more reusable CLIP-based techniques while, however, training on closed source data. In this work, we propose OpenFashionCLIP, a vision-and-language contrastive learning method that only adopts open-source fashion data stemming from diverse domains, and characterized by varying degrees of specificity. Our approach is extensively validated across several tasks and benchmarks, and experimental results highlight a significant out-of-domain generalization capability and consistent improvements over state-of-the-art methods both in terms of accuracy and recall. Source code and trained models are publicly available at: \url{https://github.com/aimagelab/open-fashion-clip}.

\end{abstract}

\section{Introduction}
\label{sec:intro}
\input{sections/01_Introduction}

\section{Related Work}
\label{sec:related}
\input{sections/02_RelatedWork}

\section{On the Adaptation of CLIP to the Fashion Domain}
\label{sec:method}
\input{sections/03_Method}

\section{Experimental Evaluation}
\label{sec:experimental_eval}
\input{sections/04_Experimental_evaluation}

\section{Conclusion}
\label{sec:conclusion}
\input{sections/05_Conclusion}

\subsubsection{Acknowledgements} This work has partially been supported by the European Commission under the PNRR-M4C2 (PE00000013) project ``FAIR - Future Artificial Intelligence Research'' and the European Horizon 2020 Programme (grant number 101004545 - ReInHerit), and by the PRIN project ``CREATIVE: CRoss-modal understanding and gEnerATIon of Visual and tExtual content'' (CUP B87G22000460001), co-funded by the Italian Ministry of University.
%
%
%
\bibliographystyle{splncs04}
\bibliography{bibliography}

\end{document}

%% file: sections/01_Introduction.tex
In the era of digital transformation, online shopping, and e-commerce have experienced an unprecedented surge in popularity. The convenience, accessibility, and variety offered by these platforms have revolutionized the way consumers engage with retail. Such digital shift creates an immense volume of data, therefore, the need for scalable and robust machine learning-based solutions to accommodate customer requirements becomes increasingly vital~\cite{shiau2020shop, zhai2019learning, zhang2018visual}. In the fashion domain, this includes tasks such as cross-modal retrieval~\cite{liu2016deepfashion, hadi2015buy}, recommendation\cite{cucurull2019context, hsiao2018creating,sarkar2023outfittransformer}, and visual product search~\cite{baldrati2021conditioned,baldrati2022conditioned, morelli2021fashionsearch,wu2021fashion,baldrati2023zero}, which play a crucial role in enhancing user experience, optimizing search functionality, and enabling efficient product recommendation systems.

To address these challenges, innovative solutions that combine vision-and-language understanding have been proposed~\cite{han2022fashionvil, zhuge2021kaleido,moratelli2023fashion}. Although prior works have made noteworthy contributions in the fashion domain, they still suffer from some deficiencies. Approaches like~\cite{baldrati2022conditioned} are able to well fit a specific task but struggle to adapt to unseen datasets and exhibit sub-optimal performance when faced with domain shifts. This results in poor zero-shot capability.

On the contrary, other techniques have employed CLIP-based methods~\cite{yaofilip,lisupervision}, which offer better generalization capabilities thanks to the pre-training on large-scale datasets. Some works as~\cite{chia2022contrastive}, have often relied on closed-source data, limiting their applicability and hindering the ability to reproduce and extend results. Therefore, there remains a need for a scalable and reusable method that can leverage open-source fashion data with varying levels of detail while demonstrating improved generalization and performance.
In response to the aforementioned challenges, in this paper, we propose \ours, a vision-and-language contrastive learning method that stands out from previous approaches in several ways. We adopt open-source fashion data from multiple sources encompassing diverse styles and levels of detail. Specifically, we adopt four publicly available datasets for the training phase, namely FashionIQ~\cite{wu2021fashion}, Fashion-Gen~\cite{rostamzadeh2018fashion}, Fashion200K~\cite{han2017automatic}, and iMaterialist~\cite{guo2019imaterialist}. We believe this approach not only enhances transparency and reproducibility but also broadens the accessibility and applicability of our technique to a wider range of users and domains.

The contrastive learning framework employed in \ours enables robust generalization capabilities, ensuring consistent performance even in the presence of domain shifts and previously unseen data. Our method adopts a fashion-specific prompt engineering technique~\cite{Radford2021LearningTV, brown2020language, gao2020making} and is able to effectively learn joint representations from multiple domains. \ours overcomes the limitations of supervised learning approaches and closed-source data training, facilitating seamless integration between visual and textual modalities.

Extensive experiments have been conducted to evaluate the effectiveness of \ours across diverse tasks and benchmarks. 
We provide a comparison against CLIP~\cite{Radford2021LearningTV}, OpenCLIP~\cite{wortsman2022robust} and a recent CLIP-based method fine-tuned on closed-source fashion data, namely FashionCLIP~\cite{chia2022contrastive}. The experimental results highlight the significant out-of-domain generalization capability of our method.
Notably, our fine-tuning strategy on open-source fashion data yields superior performance compared to competitors in several metrics, thus underscoring the benefits of leveraging open-source datasets for training.

%% file: sections/02_RelatedWork.tex
The ever-growing interest of customers in e-commerce has made 
the introduction of innovative solutions essential to enhance the online experience. 
On this basis, recommendation systems play a crucial role and numerous works have been introduced~\cite{hsiao2018creating,cucurull2019context,sarkar2023outfittransformer,de2023disentangling}. An illustrative example is the automatic creation of capsule wardrobes proposed in~\cite{hsiao2018creating}, where given an ensemble of garments and accessories the proposed method provided some possible visually compatible outfits. One of the most significant challenges of this task is the understanding of what visual compatibility means. To this aim, Cucurull~\etal~\cite{cucurull2019context} addressed the compatibility prediction problem by exploiting the context information of fashion items, whereas Sarkar~\etal~\cite{sarkar2023outfittransformer} exploited a Transformer-based architecture to learn an outfit-level token embedding which is then fed through an MLP network to predict the compatibility score. In addition, De Divitiis~\etal~\cite{de2023disentangling} introduced a more fine-grained control over the recommendations based on shape and color. 

Generally, users desire to seek a specific article in the catalog with relative ease, therefore, designing efficient multimodal systems represents another important key to success for the fashion industry. A considerable portion of user online interactions fall into the area of multimodal retrieval, the task of retrieving an image corresponding to a given textual query, and vice versa. Prior works range from more controlled environments~\cite{liu2016deepfashion, kuang2019fashion} to in-the-wild settings~\cite{hadi2015buy}, where the domain shift between query and database images is a challenging problem.

Beyond recommendations and retrieval, another research line that is currently attracting attention is the one of virtual try-on, both in 3D~\cite{santesteban2021self, santesteban2022ulnef, majithia2022robust} and 2D~\cite{morelli2022dresscode,xie2023gpvton,fincato2021viton, lee2022high,fenocchi2022dual,fincato2022transform,morelli2023ladi}. Virtual try-on aims to transfer a given in-shop garment onto a reference person while preserving the pose and the identity of the model.
A related area is the one marked by fashion image editing~\cite{pernuvs2023fice, baldrati2023multimodal, Dong_2020_CVPR}. While Dong~\etal~\cite{Dong_2020_CVPR} conditioned the fashion image manipulation process on sketches and color strokes, other approaches~\cite{pernuvs2023fice, baldrati2023multimodal} introduced for the first time a multimodal fashion image editing conditioned on text. 

Specifically, Pernu{\v{s}}~\etal~\cite{pernuvs2023fice} devised a GAN-based iterative solution to change specific characteristics of the given image based on a textual query. Baldrati~\etal~\cite{baldrati2023multimodal}, instead, focused on the creation of new garments exploiting latent diffusion models and conditioning the generation process on text, sketch, and model's pose. Solving the aforementioned downstream tasks has been made possible due to large-scale architectures explicitly trained on fashion data which effectively combine vision-and-language modalities to learn more powerful representations~\cite{zhuge2021kaleido, han2022fashionvil}.  Recent approaches exploit CLIP embeddings~\cite{Radford2021LearningTV} to obtain more scalable and robust solutions able to generalize to different domains without supervision~\cite{chia2022contrastive}, but the closed source data training represents the main flaw. 

%% file: sections/03_Method.tex
\subsection{Fashion-Oriented Contrastive Learning}
Despite the significant scaling capability of large vision-and-language models such as CLIP, such a property comes at a cost. The pre-training of these models is usually conducted on datasets that contain million~\cite{Radford2021LearningTV}, or even billion~\cite{schuhmann2022laionb} image-text pairs that, however, are gathered from the web and thus very noisy. Unfortunately, such coarse-grained annotations have been shown to lead to sub-optimal performance for vision-and-language learning~\cite{li2022blip,cornia2022universal}. Moreover, the adaptation of CLIP to the specific domain of fashion is far from trivial. Indeed, a significant part of the images contained in these datasets is associated with incomplete captions or even worse, with simple and basic tags collected exploiting posts uploaded on the web by general and non-fashion-expert users. 
Considering these flaws, an adaptation of CLIP to a specific domain, uniquely relying on a vanilla pre-trained version, would not enable the attainment of optimal results. In our context, training on fashion-specific datasets containing fine-grained descriptions of garments and fashion accessories becomes crucial to obtain powerful representations while guaranteeing generalization and robustness to solve the tasks demanded by the fashion industries.

\begin{figure}[t]
    \centering
    \includegraphics[width=0.98\textwidth]{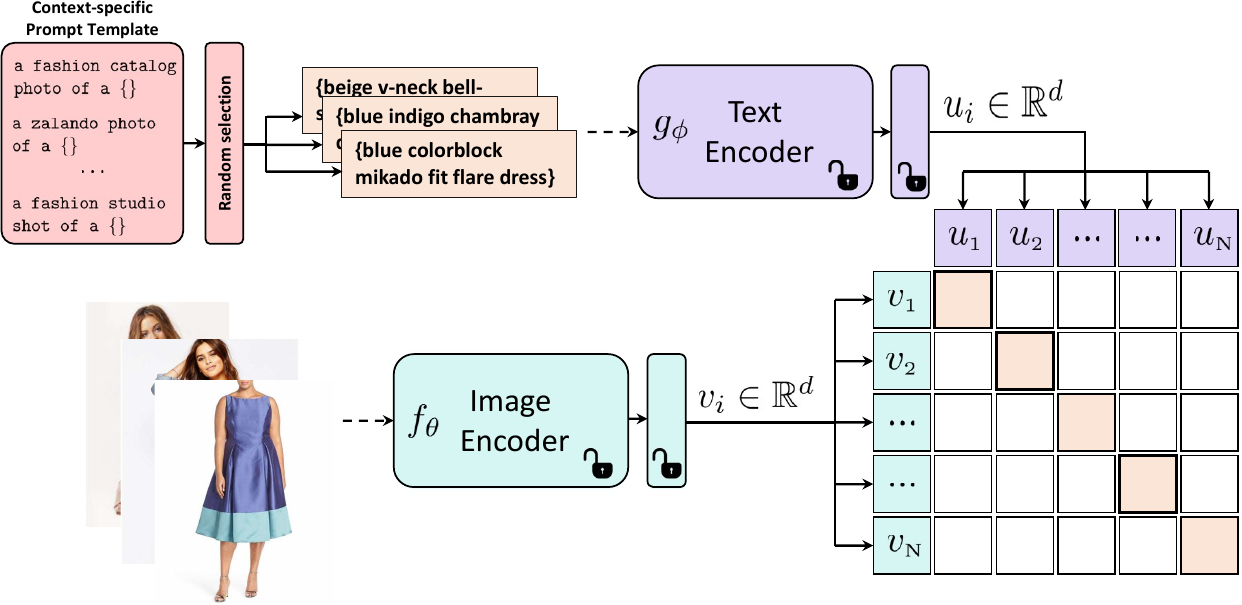}
    \vspace{-.2cm}
    \caption{Overview of our proposed method. We fine-tune both encoders and the linear projection layers toward the embedding space.}
    \label{fig:method}
\vspace{-.35cm}
\end{figure}

\subsection{CLIP Preliminaries}
Contrastive learning is a self-supervised machine learning technique that aims to learn data representations by constructing a powerful embedding space where semantically related concepts are close while dissimilar samples are pushed apart.  
On this line, the vision-and-language domain has already capitalized on such a learning technique. The CLIP model~\cite{Radford2021LearningTV} represents the most common and illustrative method for connecting images and text in a shared multimodal space. The CLIP architecture consists of a text encoder $g_\phi$ and an image encoder $f_\theta$, trained on image-caption pairs $\mathcal{S} = \{ (x_i, t_i) \}_{i=1}^N$.

The image encoder $f_\theta$ embeds an image $x \in \mathcal{X}$ obtaining a visual representation $\bm{v}=f_\theta(x)$. In the same manner, the text encoder $g_\phi$ takes as input a tokenized string $\tilde{t}$ and returns a textual embedding $\bm{u}=g_\phi(\tilde{t})$.
For each batch $\mathcal{B}$ of image-caption pairs $\mathcal{B} = \{ (x_i, t_i) \}_{i=1}^L$, where $L$ is the batch size, the objective is to maximize the cosine similarity between $\bm{v}_{i}$ and $\bm{u}_{i}$ while minimizing the cosine similarity between $\bm{v}_{i}$ and $\bm{u}_{j}$, $\forall j \neq i$. The CLIP loss can be formally expressed as the sum of two symmetric terms:
\begin{equation}
    \mathcal{L}_{contrastive} = \mathcal{L}_{T2I} + \mathcal{L}_{I2T},
\end{equation}
\begin{equation}
    \mathcal{L}_{T2I} = - \frac{1}{L} \sum_{i=1}^{L} \log \frac{\exp(\tau \bm{u}_{i}^T\bm{v}_{i})}{\sum_{j=1}^{L}\exp(\tau \bm{u}_{i}^T\bm{v}_{j})}, 
\end{equation}
\begin{equation}
    \mathcal{L}_{I2T} = - \frac{1}{L} \sum_{i=1}^{L} \log \frac{\exp(\tau \bm{v}_{i}^T\bm{u}_{i})}{\sum_{j=1}^{L}\exp(\tau \bm{v}_{i}^T\bm{u}_{j})}, 
\end{equation}
where $\tau$ represents a temperature parameter.

\subsection{Open Source Training}
In the fashion domain, several datasets, characterized by multimodal annotations from human experts, have been introduced. Differently from prior work~\cite{chia2022contrastive} that fine-tuned CLIP on a private dataset, we devise a contrastive learning strategy entirely based on open-source data. An overview of the proposed CLIP-based fine-tuning is shown in Fig.~\ref{fig:method}.
In detail, we adopt four publicly available datasets:
\tit{\textbf{Fashion-Gen}~\cite{rostamzadeh2018fashion}} The dataset contains a total of $325,536$ high resolution images ($1360 \times 1360$) with $260,480$ samples for the training set and $32,528$ images both for validation and test set. In addition, 48 main categories and 121 fine-grained categories (\ie~\emph{subcategory}) are defined.
\tit{\textbf{Fashion IQ}~\cite{wu2021fashion}} There are $77,684$ images, divided into three main categories (dresses, shirts, and tops\&tees), with product descriptions and attribute labels.
\tit{\textbf{Fashion200K}~\cite{han2017automatic}} It contains $209,544$ clothing images from five categories (dresses, tops, pants, skirts, and jackets) and an associated textual description.
\tit{\textbf{iMaterialist}~\cite{guo2019imaterialist}} It is a multi-label dataset containing over one million images and 8 groups of 228 fine-grained attributes. 

\begin{figure}[t]
    \centering
    \includegraphics[width=0.96\textwidth]{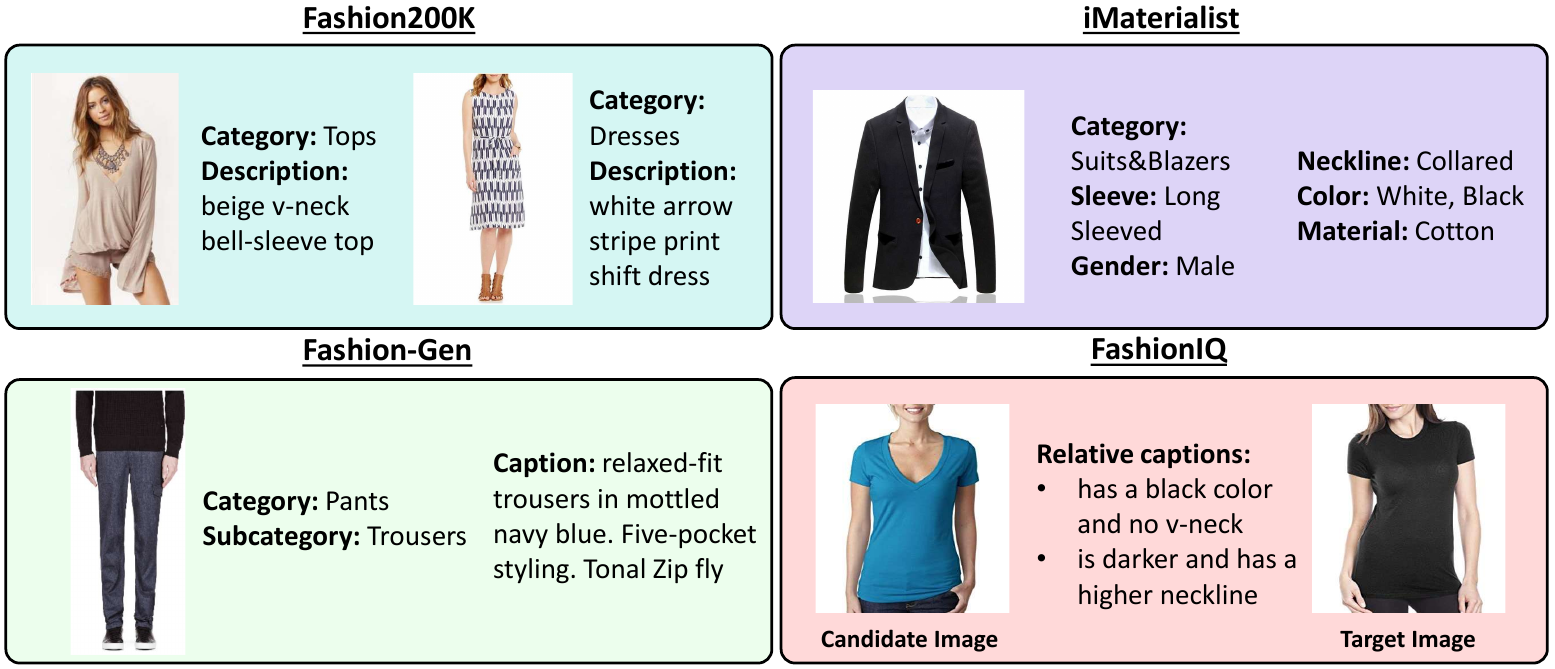}
    \vspace{-.2cm}
    \caption{Qualitative samples from the training datasets.}
    \label{fig:samples_dataset_training}
\vspace{-.35cm}
\end{figure}

These datasets are characterized by different levels of detail of the image annotations. FashionIQ has been proposed to accomplish the task of interactive image retrieval, therefore, the captions are relative to what should be modified in the source image to retrieve the target image. On the contrary, iMaterialist only contains attributes while Fashion-Gen and Fashion200K present more semantically rich descriptions. As a pre-processing step, we apply lemmatization and extract the noun chunks from the textual descriptions. Noun chunks are sequences of words that include a noun and any associated word that modifies or describes that noun (\eg~an adjective). In particular, we adopt the spaCy\footnote{\url{https://github.com/explosion/spaCy}} NLP library to extract noun chunks. Data pre-processing is performed for all datasets, except for iMaterialist which only contains simple attributes, thus making such an operation unnecessary. For the sake of clarity, from now on we refer to $t_i$ as the pre-processed caption after noun chunks extraction. Examples of image-caption pairs from the training datasets are reported in Fig.~\ref{fig:samples_dataset_training}.

Compared to FashionCLIP which was trained on approximately $700k$ images, our training set is much larger and sums to $1,147,929$ image-text pairs. In detail, during fine-tuning, we construct each batch so that it contains image-text pairs from all the different data sources. Considering the great number of pairs of the complete training dataset, we fine-tune all the pre-trained weights of the CLIP model. Indeed, only training the projections toward the embedding space would not allow to fully effectively capture the properties of the data distribution.

\subsection{Prompt Engineering}
\label{sec:prompt_engineering}
Prompt engineering is the technique related to the customization of the prompt text for each task. Providing context to the model has been shown to work well in a wide range of settings.
Following prior works~\cite{Radford2021LearningTV, brown2020language, gao2020making}, we provide our model with a fashion-specific context defining a template of prompts related to our application domain. Specifically, given a template of prompts $\mathcal{P} = \{ (p_i) \}_{i=1}^{|\mathcal{P}|}$, at each training step, we select a random $p_i \in \mathcal{P}$ for each image-caption pair $(x_i, t_i) \in \mathcal{B}$. The caption $t_i$ is concatenated to $p_i$ obtaining the final CLIP input.

The complete fashion-specific template includes the following prompts: \small{
\texttt{"a photo of a"}, \texttt{"a photo of a nice"}, 
\texttt{"a photo of a cool"}, \texttt{"a photo of an expensive"}, \texttt{"a good photo of a"}, \texttt{"a bright photo of a"}, \texttt{"a fashion studio shot of a"}, \texttt{"a fashion magazine photo of a"}, \texttt{"a fashion brochure photo of a"}, \texttt{"a fashion catalog photo of a"}, \texttt{"a fashion press photo of a"}, \texttt{"a zalando photo of a"}, \texttt{"a yoox photo of a"}, \texttt{"a yoox web image of a"}, \texttt{"an asos photo of a"}, \texttt{"a high resolution photo of a"}, \texttt{"a cropped photo of a"}, \texttt{"a close-up photo of a"}, \texttt{"a photo of one"}.}

%% file: sections/04_Experimental_evaluation.tex
In this section, we describe the open-source datasets used as benchmarks together with the tasks performed to assess the scalability and robustness of our approach.

\subsection{Benchmark Datasets}
We validate our approach across three different datasets:

\smallbreak\noindent{\textbf{DeepFashion}~\cite{liu2016deepfashion}} contains over $800,000$ images and is divided into several benchmarks.  
In our experiments, we employ the attribute prediction subset which contains $40,000$ images and $1,000$ different attributes.

\smallbreak\noindent{\textbf{Fashion-MNIST}~\cite{xiao2017fashion}} is based on the Zalando catalog and consists of $60,000$ training images, a test set of $10,000$ examples, and 10 categories. All images are in grayscale and have a $28 \times 28$ resolution. Following~\cite{chia2022contrastive}, we apply image inversion, thus working on images with a white background.

\begin{figure}[t]
    \centering
    \includegraphics[width=0.96\textwidth]{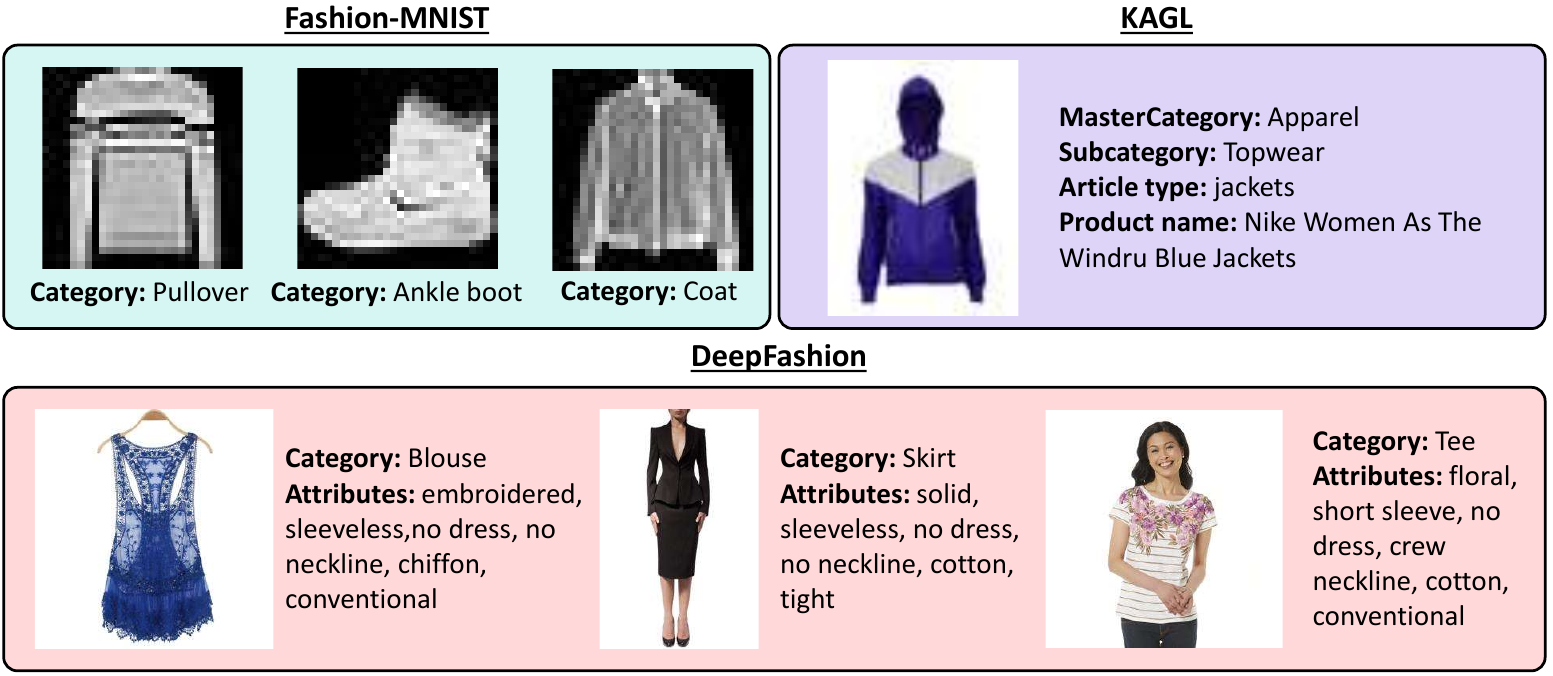}
    \vspace{-.2cm}
    \caption{Samples from the benchmark datasets.}
    \label{fig:samples_benchmark_datasets}
\vspace{-.35cm}
\end{figure}

\smallbreak\noindent{\textbf{KAGL}} is a subset of~\cite{kagl} and contains $44,441$ images equipped with textual annotations including the master category, the sub-category, the article type, and the product description. In detail, we filter out all those images not belonging to the \emph{`apparel'} master category and kept the images depicting humans, resulting in a total of $21,397$ samples, 8 sub-categories, and 58 article types. Qualitative examples of the adopted benchmarks are reported in Fig.~\ref{fig:samples_benchmark_datasets}.

\subsection{Implementation Details}
We train the final model for 60 epochs using a batch size of 2048. To save memory, we adopt the gradient checkpointing technique~\cite{chen2016training}. AdamW~\cite{loshchilovdecoupled} is employed as optimizer, with $\beta_1$ set to $0.9$ and $\beta_2$ equal to $0.98$, epsilon of $1e-6$, and weight decay equal to $0.2$. A learning rate of $5e-7$ and automatic mixed precision are applied. For a fair comparison with competitors, we select the ViT-B/32 backbone as the image encoder. During training, we apply the prompt engineering strategy described in Sec.~\ref{sec:prompt_engineering}. 

As a pre-trained CLIP model, we refer to the OpenCLIP implementation~\cite{ilharco_gabriel_2021_5143773} trained on LAION-2B~\cite{schuhmann2022laionb} composed of 2 billion image-text pairs. In the evaluation phase, following the pre-processing procedure of~\cite{Radford2021LearningTV}, we resize the image along the shortest edge and apply center crop. 

\subsection{Zero-shot Classification}
In our context, zero-shot classification refers to the task of classification on unseen datasets characterized by different data distributions compared to the training datasets. The task is crucial to assess the transfer capability of the model to adapt to new and unseen domains.  
Following the standard CLIP evaluation setup~\cite{Radford2021LearningTV}, 
we perform classification by embedding the image and all $k$ categories. Regarding prompt engineering, we always append every category \texttt{\{label\}} to the same generic prompt \texttt{"a photo of a"}. We feed each category prompt through the CLIP textual encoder $g_\phi$ obtaining a set of feature vectors $\mathcal{U} = \{(\bm{u}_i)\}_{i=1}^{N}$.
In the same manner, we feed the image $x_i$ through the CLIP image encoder to get the embedded representation $\bm{v} = f_\phi(x_i)$. To classify the image we compute the cosine similarity between $\bm{v}$ and each text representation $\bm{u}_i$. The predicted category is the one with the highest similarity. 
Experiments have been conducted on the test splits of Fashion-MNIST, KAGL, and on the attribute prediction benchmark of DeepFashion. Our model is compared against the original CLIP model~\cite{Radford2021LearningTV}, which was trained on the private WIT dataset, OpenCLIP~\cite{wortsman2022robust} pre-trained on LAION-400M~\cite{schuhmann2021laion} and LAION-2B~\cite{schuhmann2022laionb}, and FashionCLIP~\cite{chia2022contrastive} that was fine-tuned on closed source data from \emph{Farfetch}. Note that we have reproduced the results of FashionCLIP by exploiting the source code released by the authors and adapting it to our tasks and settings. The task is evaluated considering three well-known metrics, namely accuracy@$k$, recall@$k$, and weighted $F1$ score. The accuracy@$k$ computes the number of times the correct label is among the top $k$ labels predicted by the model. The recall@$k$, instead, measures the number of relevant retrieved items with respect to the total number of relevant items for a given query. The weighted F1 score accounts for the class distribution in the dataset by calculating the F1 score for each class individually and then averaging based on the class frequencies.

Quantitative results on Fashion-MNIST and KAGL are summarized in Table~\ref{tab:classification}. The first aspect to mention is the improvement against CLIP and OpenCLIP on all metrics and both datasets, indicating the effectiveness of our fine-tuning strategy enabling a strong generalization and adaptation of our model to the specific fashion domain.
Compared to FashionCLIP, our model shows better performance on Fashion-MNIST, while when tested on the 58 article types of KAGL, the results are comparable. \ours performs better with the increase of the number of considered categories.

\input{tables/cat_prediction}

\input{tables/attr_recognition}

Table~\ref{tab:attribute} shows the results on the attribute prediction benchmark of the DeepFashion dataset. Categories of this dataset are attributes describing different garment characteristics (\eg~v-neck, sleeveless, etc.), therefore we leverage the recall metric in this setting to account for the multi-label nature of the dataset. In particular, we evaluate both the per-class recall@$k$ and the overall recall@$k$ among all attributes. In this case, our solution outperforms FashionCLIP by a consistent margin, highlighting the effectiveness of our training strategy with data of different annotation detail granularity.

\input{tables/cross_modal_retrieval_kagl}

\subsection{Cross-modal Retrieval}
Cross-modal retrieval refers to the task of retrieving relevant contents from a multimodal dataset using multiple modalities such as text and images. Different modalities should be integrated to enable an effective search based on the user's input query. Cross-modal retrieval can be divided into two sub-tasks: image-to-text and text-to-image retrieval. In the first setting, given a query image $x$, we ask the model to retrieve the first $k$ product descriptions that better match the image. On the opposite, in text-to-image retrieval, given a text query, the first $k$ images that better correlate with the input query are returned. In Table~\ref{tab:cross_modal_retrieval}, we evaluate our fine-tuning method on the KAGL dataset in terms of recall@$k$ with $k=1,5,10$.
\ours performs better compared to FashionCLIP on both settings and according to all recall metrics, thus further confirming the effectiveness of our proposal.

\input{tables/ablation_prompt_engineering}

\subsection{Effectiveness of Prompt Engineering}
Finally, in Table~\ref{tab:ablation}, we evaluate the individual contribution of prompt engineering in our fine-tuning method. We present the ablation study on all considered benchmarks. The first line of the table (\ie~$w/o$ prompt engineering) refers to the case where we perform fine-tuning without using the fashion-specific template described in Sec.~\ref{sec:prompt_engineering} but employing a fixed prompt (\ie~\texttt{"a photo of a"}). Notably, Fashion-MNIST is used for the classification task, DeepFashion for retrieval, and KAGL for both. As the results demonstrate, the idea to construct a fashion-specific set of prompts clearly performs well across all cases except for the KAGL classification benchmark. We argue that in general, domain-specific prompt engineering represents a key factor to obtain greater domain adaptation of the CLIP model.

%% file: tables/cat_prediction.tex
\begin{table}[t]
\caption{Category prediction results on the Fashion-MNIST and the KAGL datasets.}\label{tab:classification}
\vspace{-.15cm}
\setlength{\tabcolsep}{.32em}
\resizebox{\linewidth}{!}{
\begin{tabular}{lcccc c cc c cccc}
\toprule
& & & & & \multicolumn{2}{c}{\textbf{F-MNIST}} & & \multicolumn{4}{c}{\textbf{KAGL}} \\
\cmidrule{6-7} \cmidrule{9-12}
\textbf{Model} & \textbf{Backbone} & \textbf{Pre-Training} & \textbf{Fine-tuned} & & Acc@1 & F1 & & Acc@1 & Acc@5 & Acc@10 & F1 \\
\midrule
CLIP & ViT-B/16 & OpenAI WIT & \xmark & & 69.29 & 67.88 & & 31.54 & 70.08 & 90.09 & 36.04 \\ 
CLIP & ViT-B/32 & OpenAI WIT & \xmark & & 69.51 & 66.56 & & 21.44 & 66.13 & 84.97 & 27.70 \\ 
OpenCLIP & ViT-B/32 & LAION-400M & \xmark & & 81.62 & 81.16 & & 33.69 & 76.60 & 89.23 & 37.89 \\ 
OpenCLIP & ViT-B/32 & LAION-2B & \xmark & & 83.69 & 82.75 & & 46.18 & 84.49 & 95.44 & 51.23 \\ 
\midrule
FashionCLIP & ViT-B/32 & LAION-2B & \cmark & & 82.23 & 82.03 & & \bf52.90 & 85.41 & 93.40 & \bf54.48 \\ 
\rowcolor{teagreen}
\textbf{\ours} & ViT-B/32 & LAION-2B & \cmark & & \bf84.33 & \bf84.19 & & 45.97 & \bf88.30 & \bf96.46 & 53.85 \\ 
\bottomrule
\end{tabular}
}
\vspace{-.1cm}
\end{table}

%% file: tables/attr_recognition.tex
\begin{table}[t]
\caption{Attribute recognition results on the DeepFashion dataset.}\label{tab:attribute}
\vspace{-.15cm}
\setlength{\tabcolsep}{.32em}
\resizebox{\linewidth}{!}{
\begin{tabular}{lcccc c ccc c ccc}
\toprule
& & & & & \multicolumn{3}{c}{\textbf{Overall Recall}} & & \multicolumn{3}{c}{\textbf{Per-Class Recall}} \\
\cmidrule{6-8} \cmidrule{10-12}
\textbf{Model} & \textbf{Backbone} & \textbf{Pre-Training} & \textbf{Fine-tuned} & & R@3 & R@5 & R@10 & & R@3 & R@5 & R@10 \\
\midrule
CLIP & ViT-B/16 & OpenAI WIT & \xmark & & 8.00 & 11.40 & 17.54 & & 13.31 & 17.42 & 24.54 \\ 
CLIP & ViT-B/32 & OpenAI WIT & \xmark & & 7.35 & 10.30 & 16.60 & & 11.39 & 15.13 & 21.67 \\ 
OpenCLIP & ViT-B/32 & LAION-400M & \xmark & & 12.58 & 17.22 & 25.64 & & 17.9 & 22.81 & 30.71 \\ 
OpenCLIP & ViT-B/32 & LAION-2B & \xmark & & 13.07 & 17.70 & 26.13 & & 19.35 & 24.31 & 32.51 \\ 
\midrule
FashionCLIP & ViT-B/32 & LAION-2B & \cmark & & 15.19 & 20.83 & 32.37 & & 17.30 & 22.27 & 30.56 \\ 
\rowcolor{teagreen}
\textbf{\ours} & ViT-B/32 & LAION-2B & \cmark & & \bf24.47 & \bf32.97 & \bf45.77 & & \bf28.67 & \bf36.07 & \bf47.28 \\ 
\bottomrule
\end{tabular}
}
\vspace{-.4cm}
\end{table}

%% file: tables/cross_modal_retrieval_kagl.tex
\begin{table}[t]
\caption{Cross-modal retrieval results on the KAGL dataset.}\label{tab:cross_modal_retrieval}
\vspace{-.15cm}
\setlength{\tabcolsep}{.32em}
\resizebox{\linewidth}{!}{
\begin{tabular}{lcccc c ccc c ccc}
\toprule
& & & & & \multicolumn{3}{c}{\textbf{Image-to-Text}} & & \multicolumn{3}{c}{\textbf{Text-to-Image}} \\
\cmidrule{6-8} \cmidrule{10-12}
\textbf{Model} & \textbf{Backbone} & \textbf{Pre-Training} & \textbf{Fine-tuned} & & R@1 & R@5 & R@10 & & R@1 & R@5 & R@10 \\
\midrule
FashionCLIP & ViT-B/32 & LAION-2B & \cmark & & 6.61 & 19.23 & 28.66 & & 6.97 & 19.14 & 27.49 \\ 
\rowcolor{teagreen}
\textbf{\ours} & ViT-B/32 & LAION-2B & \cmark & & \bf7.57 & \bf20.72 & \bf30.38 & & \bf7.73 & \bf20.58 & \bf28.56 \\ 
\bottomrule
\end{tabular}
}
\vspace{-.1cm}
\end{table}

%% file: tables/ablation_prompt_engineering.tex
\begin{table}[t]
\caption{Ablation study to assess the validity of the prompt engineering technique.}\label{tab:ablation}
\vspace{-.15cm}
\setlength{\tabcolsep}{.35em}
\resizebox{\linewidth}{!}{
\begin{tabular}{lc cc c cc c cc c cc}
\toprule
& & \multicolumn{2}{c}{\textbf{F-MNIST}} & & \multicolumn{2}{c}{\textbf{KAGL}} & & \multicolumn{2}{c}{\textbf{DeepFashion}} & & \multicolumn{2}{c}{\textbf{KAGL}} \\
\cmidrule{3-4} \cmidrule{6-7} \cmidrule{9-10} \cmidrule{12-13}
\textbf{Model} & & Acc@1 & F1 & & Acc@1 & F1 & & R@3 & R@3 (cls) & & R@1 (I2T) & R@1 (T2I) \\
\midrule
w/o prompt engineering & & 83.21 & 82.99 & & \bf47.51 & 47.3 & & 20.34 & 25.21 & & 7.47 & \bf7.73 \\ 
\rowcolor{teagreen}
\textbf{\ours} & & \bf84.33 & \bf84.19 & & 45.97 & \bf53.85 & & \bf24.47 & \bf28.67 & & \bf7.57 & \bf7.73 \\ 
\bottomrule
\end{tabular}
}
\vspace{-.4cm}
\end{table}

%% file: sections/05_Conclusion.tex
In this paper, we introduced \ours, a vision-and-language contrastive learning method designed to address the scalability and robustness challenges posed by the fashion industry for online shopping and e-commerce. By leveraging open-source fashion data from diverse sources, \ours overcomes limitations associated with closed-source datasets and enhances transparency, reproducibility, and accessibility. Our strategy, characterized by the fine-tuning of all pre-trained weights across all CLIP layers together with the adoption of a context-specific prompt engineering technique, effectively enables better adaption to our specific domain. We evaluated our strategy on three benchmarks and demonstrated that the proposed solution led to superior performance over the baselines and competitors achieving better accuracy and recall in almost all settings. 